\documentclass[sigconf]{acmart}

\AtBeginDocument{%
  \providecommand\BibTeX{{%
    \normalfont B\kern-0.5em{\scshape i\kern-0.25em b}\kern-0.8em\TeX}}}

\usepackage{amsmath,array,graphicx}
\usepackage{kantlipsum}

\setcopyright{acmcopyright}

\copyrightyear{2023}
\acmYear{2023}
\setcopyright{acmlicensed}\acmConference[MM '23]{Proceedings of the 31st
ACM International Conference on Multimedia}{October 29-November 3,
2023}{Ottawa, ON, Canada}
\acmBooktitle{Proceedings of the 31st ACM International Conference on
Multimedia (MM '23), October 29-November 3, 2023, Ottawa, ON, Canada}
\acmPrice{15.00}
\acmDOI{10.1145/3581783.3612858}
\acmISBN{979-8-4007-0108-5/23/10}

\begin{document}

\title{MAGIC-TBR: Multiview Attention Fusion for Transformer-based Bodily Behavior Recognition in Group Settings}

\author{Surbhi Madan}
\orcid{0009-0000-3774-8117}
\affiliation{%
 \institution{Indian Institute of Technology Ropar}
  \country{India}
}

\author{Rishabh Jain}
\orcid{0009-0000-1021-5038}
\affiliation{%
 \institution{Indian Institute of Technology Ropar}
  \country{India}
}

\author{Gulshan Sharma}
\orcid{0000-0002-5332-7256}
\affiliation{%
 \institution{Indian Institute of Technology Ropar}
  \country{India}
}

\author{Ramanathan Subramanian}
\orcid{0000-0001-9441-7074}
\affiliation{%
 \institution{University of Canberra}
 \country{Australia}
}

\author{Abhinav Dhall}
\orcid{0000-0002-2230-1440}
\affiliation{%
 \institution{Indian Institute of Technology Ropar}
 \country{India}
 \institution{\& Monash University}
 \country{Australia}
}
\settopmatter{printacmref=true}

\begin{abstract}
Bodily behavioral language is an important social cue, and its automated analysis helps in enhancing the understanding of artificial intelligence systems. Furthermore, behavioral language cues are essential for active engagement in social agent-based user interactions. Despite the progress made in computer vision for tasks like head and body pose estimation, there is still a need to explore the detection of finer behaviors such as gesturing, grooming, or fumbling. This paper proposes a multiview attention fusion method named MAGIC-TBR that combines features extracted from videos and their corresponding Discrete Cosine Transform coefficients via a transformer-based approach. The experiments are conducted on the BBSI dataset and the results demonstrate the effectiveness of the proposed feature fusion with multiview attention.
The code is available at: \url{https://github.com/surbhimadan92/MAGIC-TBR}
\end{abstract}



%\ccsdesc[500]{Human-centered computing~Empirical studies in HCI}


\maketitle

\section{Introduction and Background}

Recognizing human behavior allows for intuitive and natural interactions with the technology. By understanding behavioral cues, computer systems can respond more appropriately, which improves the user experience and makes human-computer interaction more engaging \cite{madan2021head}. These behavior cues may include a wide range of observable actions such as body language, speech patterns, and microexpressions. Collectively, these behavior cues contribute to our understanding of human interaction, allowing us to respond effectively in various social contexts.

Body language is a powerful social cue which greatly influence how others perceive and interpret our communication \cite{mandal2014nonverbal,Jagan18,Sub13,Lepri10}. It contains non-verbal signals such as facial expressions, gestures, and body posture. These cues provide important information about our emotions \cite{joshi2013can} and convey information which words alone cannot express \cite{L2}. The automatic analysis of body language is widely studied in the context of human-computer interaction \cite{lan2021attention}. By understanding the dynamics of body language, one can interpret the underlying emotions, intentions, and attitudes of users. An example which illustrates the correlation of body language and verbal communication is when someone involuntarily smiles upon receiving a pleasant news. This serves as evidence of how body language aids in understanding emotions, in addition to verbal cues \cite{L5}.

\begin{figure*}[!tbph]
    \centering
    \includegraphics[scale = 0.58]{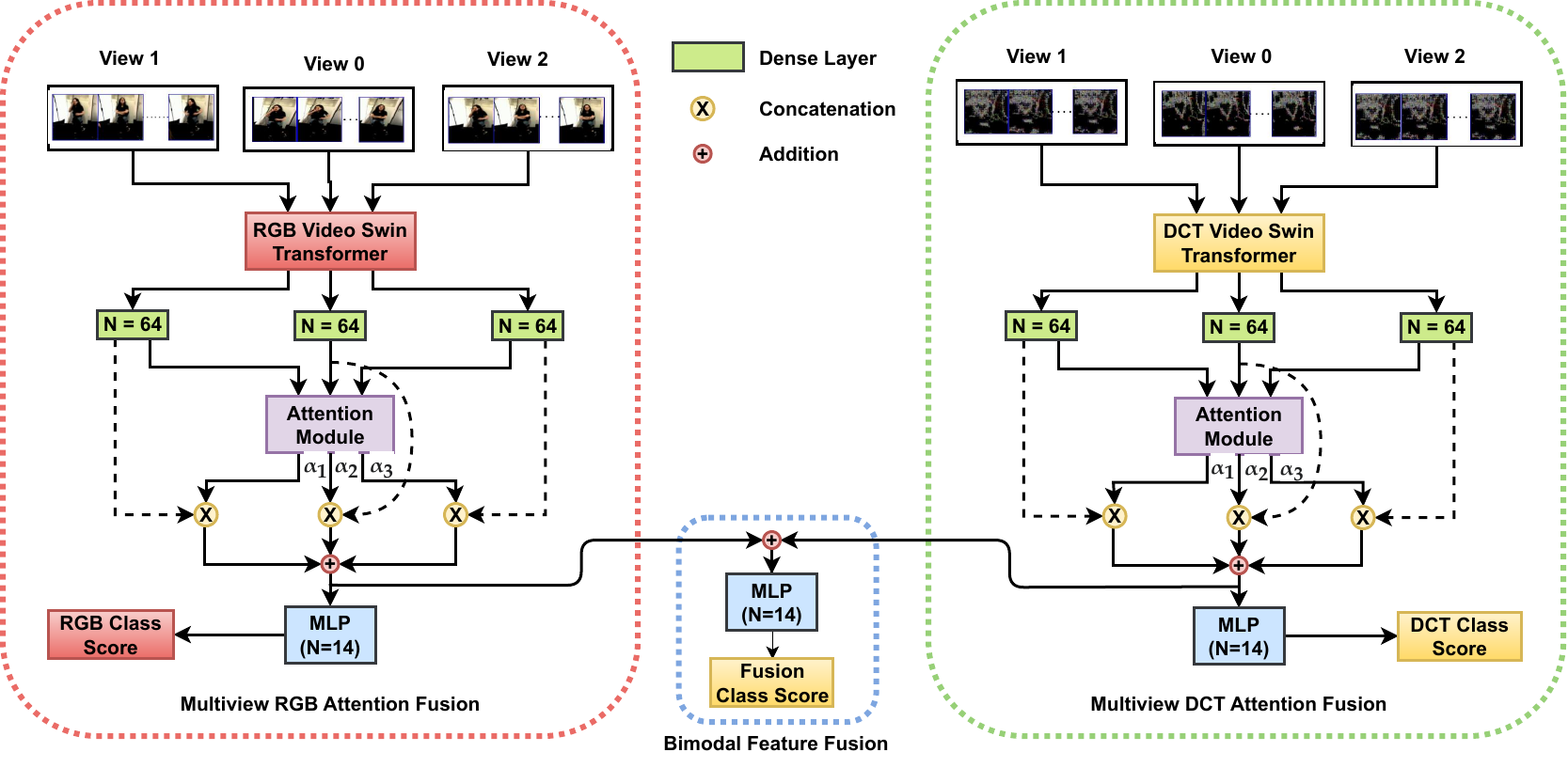}
    \caption{Overall architecture of MAGIC-TBR. The method is based on bimodal fusion of RGB and DCT features.}
    \label{main}
\end{figure*}

Although body language can offer valuable insights into user's emotions, there are some limitations when it comes to the interpretation part. Body language doesn't follow strict grammatical rules, and it needs to be subjectively interpreted as body movements don't always have a definite meaning \cite{L7}. Moreover, it is crucial to consider interpersonal differences, as the same body language may hold different meanings across different individuals. Contextual factors further complicate the interpretation of body language as the same individual may exhibit different body language in different social settings.

This paper we focus on recognizing bodily behavior through videos during a constrained social interaction scenario. Our objective is to understand an individual's body behavior in a group interaction setting when 3-4 individuals discuss a controversial topics. To this end, we propose a Multiview Attention Fusion for Transformer-based Bodily Behavior Recognition (MAGIC-TBR), a simple yet effective approach which captures discriminative and complementary feature representation from either view. The primary contribution of our work lies in integrating Swin Transformer-based RGB and DCT features which effectively combines the spatial and motion information for multi-view fusion. This integration leads to the creation of robust video-level features. We demonstrate through extensive experiments on the benchmark dataset \cite{balazia2022bodily} that our MAGIC-TBR approach improves bodily behavior recognition performance compared to the baseline method \cite{baseline}.

% \section{Experimental Details}

% \subsection{Prediction Settings} The problem is framed as a multi-label classification with 14 classes. To address the imbalanced class distribution, we apply macro-averaged average precision for evaluation, which computes the average precision per class and aggregates them using macro averaging. During training, early stopping with a patience of 5 epochs is implemented to prevent model degradation on validation sets.

\section{Method}

\subsection{Problem Formulation}
Given a 64-frame video snippet (V) of a person $p_i$ interacting with a group of people P ($p_i \in$ P), where the video captures the person's seated body and face without audio. The objective is to predict the likelihood of 14 behavior classes present in the video using multilabel classification. Additionally, two side-view videos from both the left and right-hand perspectives are available. 

\subsection{Dataset}
The BBSI dataset \cite{balazia2022bodily} provides annotations on the MPIIGroupInteraction dataset \cite{muller2018detecting}, comprising 22 group discussions, each participant engaging in 20-minute discussions on controversial topics. The BBSI dataset annotates 15 bodily behavior classes including \emph{gesture}, \emph{fumble}, \emph{hand-face}, \emph{hand-mouth}, and \emph{legs-crossed}. It contains 2.87 million annotated frames, capturing 26 hours of human behavior during continuous group interactions. Interested readers may refer to \cite{balazia2022bodily, muller2018detecting} for more information on the dataset. In order to ensure consistency in the dataset, we resize the original videos to $224\times224$. Additionally, we exclude videos which have fewer than 64 frames, as they may not provide sufficient information for analysis. Furthermore, we remove videos which contain occlusions and missing information in either of the multiview perspectives. 

\subsection{Features Extraction}
We apply the following feature extraction methods:
\subsubsection{Discrete Cosine Transform (DCT)}
\begin{figure}[t]
     %\begin{minipage}{.49\textwidth}
      \centering
\includegraphics[width = \linewidth]{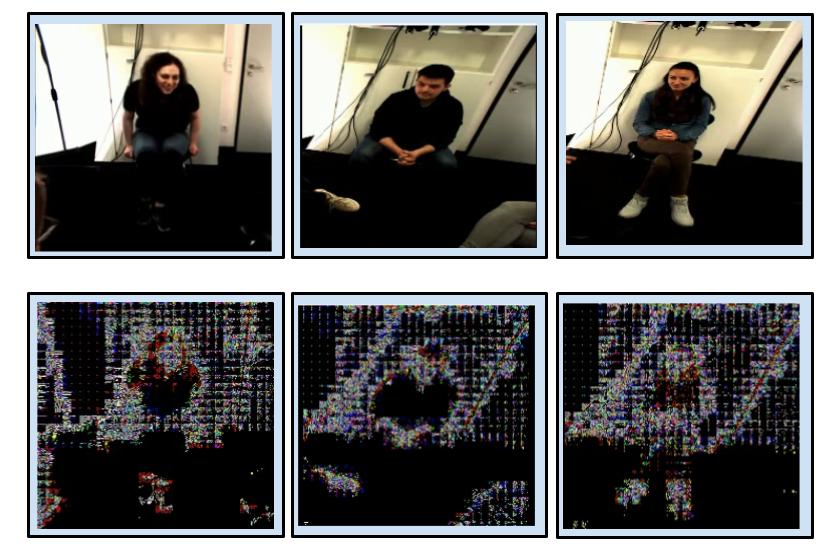}
\vspace{-3mm}
    \caption{Top: RGB frames from input videos and Bottom: corresponding DCT frames. DCT helps in finding the areas likely to contain edges or boundaries.} \label{fig: frames}\vspace{-2mm}
\end{figure}

DCT represents the image content in frequency domain as a sum of cosine functions of different frequencies and amplitudes \cite{DCT}. High-frequency DCT coefficients capture the transition in pixel intensities across small spatial regions including edges and textures. 
Frequency domain representations allows to capture specific image features and properties that is not directly observable in pixel values. We apply DCT to every RGB frames and recombine these transformed frames to generate a DCT video.  Figure \ref{fig: frames} displays examples of RGB and their corresponding DCT frames.

\subsubsection{Video Swin Transformer} 
Swin transformer \cite{liu2022video} contains a hierarchical structure with shifted windows to capture visual information efficiently. This Transformer incorporates a spatial-temporal attention mechanism, enabling it to learn complex visual features dynamically. We fine-tune the pre-trained video swin transformer's weights on both RGB \& DCT using 32-frame videos from the BBSI dataset. For implementation, we have used the MMaction2 toolbox \cite{mmaction2}, and the fine-tuned swin RGB and swin DCT networks serve as feature extractors, producing 1024-dimensional feature vectors.

\subsubsection{LaViLa Vision Features}
Video-language networks provide contextual information, which can draw attention to specific events in the input videos. Motivated by this, we have applied the LaViLa \cite{zhao2023learning} which learns video-language representations via a large language model and generates textual descriptions of the video clips. We apply the basic LaViLa network to extract video features. This network randomly sample four frames per video clip and encode their features into a ($256\times768$) dimensional vector.

%The LaViLa framework learns video-language representations via a large language model and generates textual descriptions of the video clips  \cite{zhao2023learning}. We apply LaViLa to extract video features from randomly sampled 4 frames per video, resulting in a ($256\times768$) dimensional feature vector.

\subsection{Classification Methods}
We apply following classification methods to calculate the likelihood of 14 behavior classes.

\subsubsection{Multiview Attention Fusion Method}
We apply attention-based multiview fusion, as described in \cite{sharma2018multichannel,madan2023}, to assign importance to the three views of each person. We construct two separate networks, namely \textit{Multiview RGB Attention Fusion}  (Figure \ref{main}: left) and \textit{Multiview DCT Attention Fusion} (Figure \ref{main}: right), respectively. These networks have similar configurations and generates class scores for RGB and DCT videos. \\

The multiview attention model takes swin-transformer-generated view-specific features as input. These inputs are processed through respective dense layers with 64 neurons, resulting in fixed-length feature vectors. These view descriptors are then concatenated and passed through a fully connected layer, followed by a softmax layer with three neurons for calculating attention scores (Attention Module). Generated attention scores are utilized to determine the relative importance of each view. We also apply layer normalization to generated feature vectors. To fuse the normalized features, an additive layer is employed to sum the weighted view features and process them through a dense layer for multilabel classification using a sigmoidal activation function..

% Our Multiview Attention fusion method inputs given 3 views of the video and pass them into independent swin transformers and dense layers to generate the fixed-length view descriptors. These view descriptors are then input into a common attention module which generates attention scores determining the relative importance of the each view. We further apply a dense layer with sigmoid activation to estimate the multilabel classification score.

\subsubsection{Bimodal Feature Fusion:} The Bimodal Feature Fusion approtch aims to combine the feature vectors generated by the RGB Multiview Attention Fusion (Figure \ref{main}: left and the DCT Multiview Attention Fusion (Figure \ref{main}: right). To achieve this, we apply a single addition layer followed by a dense layer with fourteen neurons (each corresponding to a specific class) and a sigmoid activation function for the multilabel classification. The overall network is depicted in Figure \ref{main}. 

\subsubsection{Transformer on LaViLa} 
We apply a transformer network inspired from  \cite{sharma_g} on the video features extracted from the LaViLa framework. As we are dealing with a multi-label classification problem, we apply only the encoder head comprising multi-head self-attention, a feed-forward network, and a classification head. 

\subsubsection{Trimodal Feature Fusion:}
The Trimodal Feature Fusion is an extension of the bimodal feature fusion approach which includes 768-dimensional LaViLa transformer generated features as a third modality.

    \begin{table}[!htbp]
    \centering
  \caption{Comparison of overall mean average precision (MAP) scores on the validation and test sets for different methods. }
  \label{tab1}
  \resizebox{\columnwidth}{!}{
 \begin{tabular}{lcc}

    \toprule
    \textbf{Methods}                  & \textbf{Validation MAP} & \textbf{Test MAP} \\
    \midrule

\textbf{Baseline}                                  & 0.41           & 0.56      \\
\textbf{Transformer on Lavila}                     & 0.25           & -        \\
\textbf{Multiview DCT}                             & 0.35           & -        \\
\textbf{Multiview RGB}                             & 0.45           & -        \\
\textbf{Trimodal Fusion: RGB $+$ DCT $+$ Transformer} & 0.47           & -        \\
\textbf{Bimodal Fusion: RGB $+$ DCT}               & \textbf{0.49 }          & \textbf{0.57}   \\
    \bottomrule
  \end{tabular}
  }
\end{table}

\begin{table*} [h!]
\centering
\fontsize{7.5}{7.5}\selectfont
	\renewcommand{\arraystretch}{1.2}
   %\fontsize{4}{4}
  \caption{Comparison of classwise mean average precision (MAP) scores on the validation set for different methods. AC: Adjusting clothing, LM: Leg Movements, Transformer: Transformer Trained on LaVila Features, Trimodal Fusion: Lavila+RGB+DCT while Bimodal: RGB+DCT.}
  \label{tab2}
    % \resizebox{18 cm}{!}{
    \scalebox{0.95}{
  \begin{tabular}{lllllllllllllll}
  %\begin{tabular}{ccccccccccccccc}
    \toprule

%\begin{tabular}{lllllllllllllll}
\textbf{Methods}                                            & \begin{tabular}[c]{@{}l@{}}\textbf{Hand}\\\textbf{Face}\end{tabular} & \begin{tabular}[c]{@{}l@{}}\textbf{Hand}\\\textbf{Mouth}\end{tabular} & \textbf{Gesture} & \textbf{Fumble} & \textbf{Scratch} & \textbf{Streching} & \begin{tabular}[c]{@{}l@{}}\textbf{Smearing}\\\textbf{Hands}\end{tabular} & \textbf{Shrug} & \begin{tabular}[c]{@{}l@{}}\textbf{AC}\end{tabular} & \textbf{Groom} & \begin{tabular}[c]{@{}l@{}}\textbf{Fold}\\\textbf{Arms}\end{tabular} & \begin{tabular}[c]{@{}l@{}}\textbf{LM}\end{tabular} & \textbf{Settle} & \begin{tabular}[c]{@{}l@{}}\textbf{Legs }\\\textbf{Crossed}\end{tabular}  \\

\midrule
\textbf{Baseline \cite{baseline}}                                  & 0.511                                                                 & 0.431                                                                  & 0.82             & 0.377           & 0.122            & 0.004              & 0.07                                                                       & 0.044          & 0.239                                                                          & 0.582          & \textbf{0.958}                                                        & 0.295                                                                    & 0.335           & \textbf{0.949}                                                            \\
\textbf{Transformer}                     & 0.562                                                       & 0.482                                                        & 0.526            & 0.279           & 0.062            & 0.003              & 0.017                                                                      & 0.013          & 0.046                                                                          & 0.429          & 0.156                                                       & 0.042                                                                    & 0.046           &0.806                                                          \\
\textbf{Multiview DCT}                             & 0.437                                                                 & 0.264                                                                  & 0.719            & \textbf{0.508}  & 0.132            & \textbf{0.009}     & 0.044                                                                      & 0.031          & 0.134                                                                          & 0.404          & 0.558                                                                 & \textbf{0.405}                                                           & 0.374           & 0.854                                                                     \\
\textbf{Multiview RGB}                             & 0.717                                                                 & 0.51                                                                   & 0.835            & 0.469           & 0.152            & 0.004              & 0.097                                                                      & 0.09           & \textbf{0.35}                                                                  & 0.618          & 0.934                                                                 & 0.263                                                                    & 0.407           & 0.913                                                                     \\
\textbf{Trimodal Fusion}  & 0.711                                                                 & 0.534                                                                  & 0.86            & 0.491           & \textbf{0.298}            & 0.004              & \textbf{0.136}                                                                      & 0.048          & 0.174                                                                          & 0.636          & 0.914                                                                 & 0.351                                                                    & 0.392           & 0.902                                                                     \\
\textbf{Bimodal Fusion}   & \textbf{0.788  }                                                               & \textbf{0.551}                                                                  & \textbf{0.871}            & 0.497           & 0.221            & 0.006              & 0.129                                                                      & \textbf{0.097}          & 0.344                                                                          & \textbf{0.691}          & 0.927                                                                 & 0.398                                                                    & \textbf{0.408}        & 0.904          \\                                                \bottomrule    
\end{tabular}
}
\end{table*}

% \textbf{Baseline} We compare the proposed approach with the challenge baseline method \cite{baseline}, which applies a pre-trained video swin transformer \cite{liu2022video} to extract features from each view. The baseline method calculates scores by aggregating the mean or maximum scores predicted by each view. The evaluation of the results demonstrates that the best performance is achieved when incorporating the background class and calculating the mean of each view's score. The inclusion of the background class involves assigning label 15 to samples that do not have any class assigned to them. Results of baseline are shown in the first row of Table \ref{tab1} and Table \ref{tab2}.

%\subsection{Prediction Type} The Problem is formulated as a 14-class multi-label classification. Early stopping with a patience value of 5 epochs is employed to prevent model degradation.\subsection{Performance  Metrics} To counter the imbalanced class distribution in multilabel classification, To counter class imbalances, we use macro averaged average precision for evaluation (i.e. computed average precision per class and aggregated using macro averaging, i.e. giving the same weight to each class.

\section{Results and Discussion}
In this section, we present the outcomes of our experiments and compare with the baseline method \cite{baseline}. Given the presence of imbalanced class samples in this 14-class multi-label classification problem, we apply mean average precision as the evaluation metric.

\subsection{Experimental Details}
We perform our experiments on a Nvidia A100 GPU, equipped with 40 GB GPU memory. When training the transformer on LaviLa features, we apply binary crossentropy loss and train the network with the Adam optimizer at a learning rate of 0.001.  We observe that for Multiview attention fusion, Bimodal, and Trimodal fusion methods, stochastic gradient descent at learning rate of 0.01 yields the best results in reducing the binary crossentropy loss. We set the number of epochs to 300 and apply early stopping with a parameter set at 10.

\subsection{Overall Results}
We present overall classification results on the validation and test sets in Table \ref{tab1}. Our proposed methods exhibit better performance than the baseline approach on the validation set, with the exception of the Transformer on Lavila and Multiview DCT. Multi-view RGB and Trimodal fusion outperforms the baseline MAP score by 0.04 and 0.06 points respectively. The bimodal fusion technique, which involves the early fusion of multiview RGB and DCT generated feature vectors, surpasses all other methods in terms of MAP score. It achieves a validation MAP of 0.49 and a test MAP of 0.57.

\subsection{Classwise Results}
We present classwise classification results on the validation sets in Table \ref{tab2}. We observe that our classification methods perform better in terms of overall MAP score, but while we carefully analyse classwise MAP scores we observe \emph{legs-crossed} and \emph{fold-arm} are better classified via the baseline method.

\subsubsection{The Multiview DCT}
We observe Multiview DCT method surpasses the baseline performance in five specific classes: \emph{fumble}, \emph{scratching}, \emph{stretching}, \emph{leg movements}, and even \emph{settle}. Except for \emph{leg movement} and \emph{settle}, all classes involve hand movements. A previous study \cite{R1} indicate that hand trajectory signals result in the increase in the high-frequency components. We observe, for \emph{fumble}, \emph{leg movement}, and \emph{stretching} class this method outperforms all other methods due to the predominance of high-frequency components.

\subsubsection{The Multiview RGB} 
We observe that Multiview RGB method outperforms the baseline, with the exception of the \emph{fold-arms} and \emph{legs-crossed} class. This improvement can be attributed to the robust representation obtained by incorporating information from all the views. In many cases, the RGB frontal view may have limited visibility, particularly around legs. However, side views provide more lighting and a left or right profile, allowing the model to capture additional information.

\subsubsection{Transformers on LaViLa} 
We observe Transformer trained on LaViLa features works better for \emph{hand-on-face} and \emph{hand-on-mouth} classes in comparison with baseline method. These classes involve static postures and distinct facial characteristics. LaViLa vision features, derived from the TimeSformer \cite{tsf} Transformer, effectively capture the temporal dependencies and interactions between frames over time. In contrast, other classes exhibit relatively lower recognition accuracy. The reason behind this could be that the Lavila model we applied extracts features from only a random selection of 4 frames, which could potentially lead to a loss of motion information.

\subsubsection{Bimodal \& Trimodal Fusion} 
We observe bimodal fusion integrating robust representations from multiview DCT and RGB, exhibits outstanding performance among other methods.
Additionally, the trimodal fusion approach (RGB+DCT+LaViLa) achieves comparable results to bimodal for most classes.

%\vspace{-4mm}

\subsection{Initial Explainability}
\begin{figure}[t]
\centering
\includegraphics[scale=0.17]{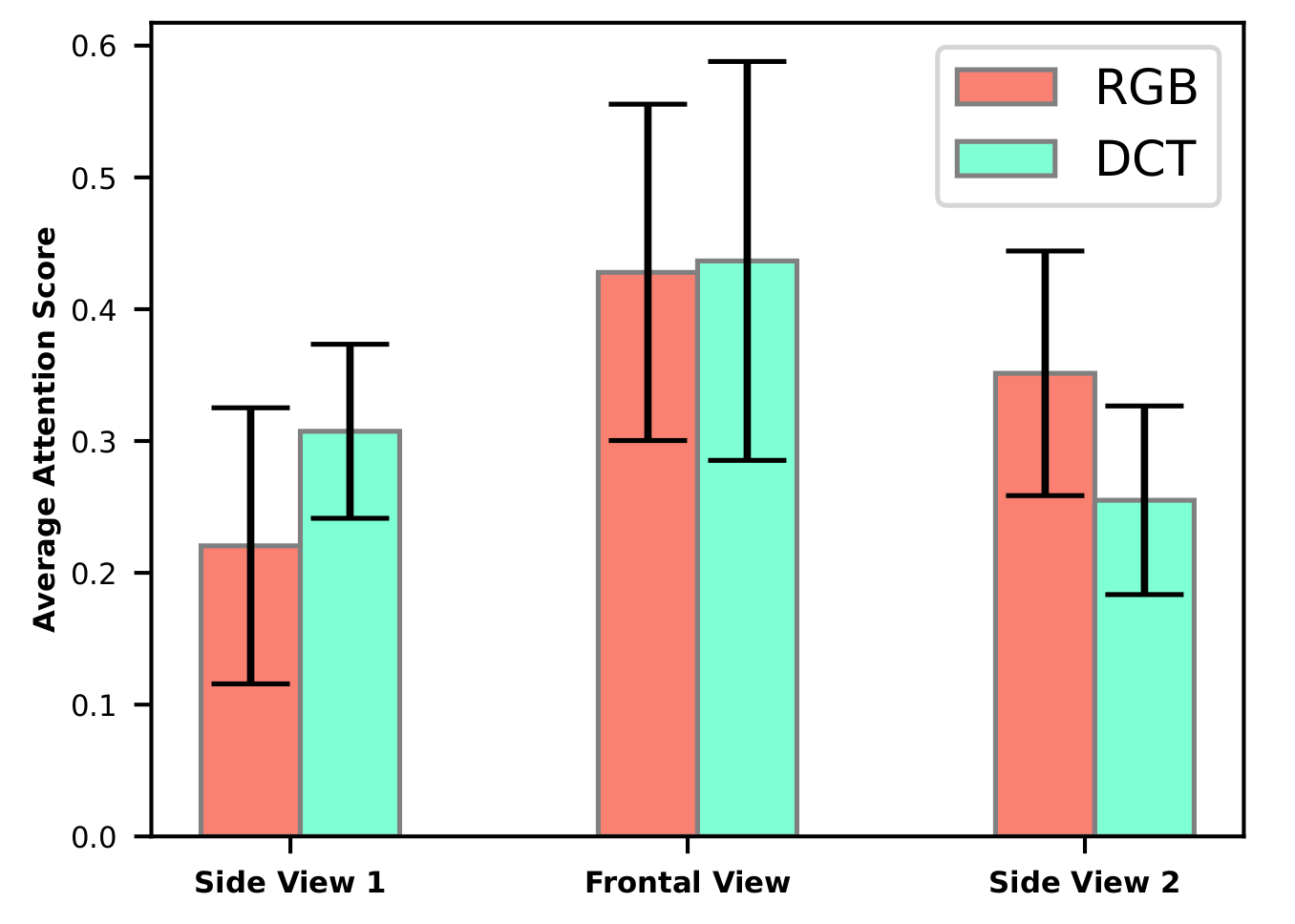}\vspace{-3mm}
\caption{The average attention scores on the validation set. } \label{fig:att}
\vspace{-3 mm}
\end{figure}

To achieve initial explanations, we analyze attention scores from RGB and DCT Multiview models in bimodal fusion (Figure \ref{fig:att}). During our analysis of the dataset videos, we observe that view 2 has lower illumination compared to view 1, especially in subject position 4. Additionally, there is occlusion in the leg areas of View 1 for subject position 2 due to the close placement of sitting chairs. The attention scores highlight the frontal view as having the highest attention, offering focused and less occluded videos for both RGB and DCT. DCT prioritizes view 1, focusing on highly illuminated videos, as DCT is more efficient for illumination variation \cite{mustafa2019image}, resulting in more accurate high-frequency components, while RGB emphasizes view 2, avoiding occlusion. Overall, multiview attention fusion contributes to good classification performance. However, this explanation is limited to treating all classes collectively, with future work focusing on classwise explanations.

\section{Conclusion}
In this paper, we present MAGIC-TBR approach which recognizes 14 bodily behavioral classes from multiview input videos. We have incorporated these multiview (RGB and DCT) networks into bimodal and trimodal fusion settings and evaluated the transformer model's performance on LaViLa features. Our experimental results indicate that classes involving hand movements such as \emph{fumble}, \emph{streching}, and \emph{scratching} are better recognized by Multiview DCT due to the presence of high-frequency components. In contrast, classes involving distinct facial characteristics such as \emph{hand-face} and \emph{hand-mouth} are better recognizable by transformer on Lavila, which preserves interactions between frames. Bimodal Fusion is outperforming all other methods, including trimodal, possibly due to the need for a more robust representation of LaViLa features for fusion. While comparing bimodal fusion with baseline, we observe an increment of 0.08 and 0.01 in validation and test MAP, respectively. We have also attempted to  provide initial explainability via generated attention scores. 

For future work, we propose to focus on enhancing the prediction performance of classes with subtle movements and limited samples such as \emph{smearing hands} and \emph{scratching}. Furthermore, incorporating advanced LaViLa-based features and building a more robust fusion architecture can also be explored.

\bibliographystyle{ACM-Reference-Format}
\bibliography{references}

\end{document}